\documentstyle[aclap]{article}

\title{Building a Generation Knowledge Source using Internet-Accessible
Newswire  }

\author{
        Dragomir R. Radev and Kathleen R. McKeown \\
        Department of Computer Science \\
        Columbia University \\
        New York, NY 10027 \\
        \{radev,kathy\}@cs.columbia.edu \\
}

\input{epsf}

\begin{document}
\maketitle

\begin{abstract}
In this paper, we describe a method for automatic creation of a knowledge
source for text generation using information extraction over the
Internet. We present a prototype system called PROFILE which uses a
client-server architecture to extract noun-phrase descriptions of entities
such as people, places, and organizations. The system serves two
purposes: as an information extraction tool, it allows users to search for
textual descriptions of entities; as a utility to generate functional
descriptions (FD), it is used in a functional-unification based generation
system. We present an evaluation of the approach and its applications to
natural language generation and summarization.  
\end{abstract}

%%%%%%%%%%%%%%%%%%%%%%%%%%%%%%%%%%%%%%%%%%%%%%%%%%%%%%%%%%%%%%%%%%%%%%%%%%%
%SECTION 1
%%%%%%%%%%%%%%%%%%%%%%%%%%%%%%%%%%%%%%%%%%%%%%%%%%%%%%%%%%%%%%%%%%%%%%%%%%%
\section{Introduction}

In our work to date on news summarization at Columbia
University \cite{McKeown&Radev95,Radev96}, information is extracted from a
series of input news articles 
\cite{muc,grishman&al92} and is analyzed by a generation component to
produce a summary that shows how perception of the event has changed
over time. In this summarization paradigm, problems arise when
information needed for the summary is either missing from the input
article(s) or not extracted by the information extraction system. In
such cases, the information may be readily available in other current
news stories, in past news, or in online databases. If the
summarization system can find the needed information in other online
sources, then it can produce an improved summary by merging
information from multiple sources with information extracted from the
input articles.

In the news domain, a summary needs to refer to 
people, places, and organizations and provide descriptions that clearly
identify the entity for  the reader. Such descriptions may 
not be present in the original text that is being summarized. 
For example, the American pilot Scott O'Grady, downed in Bosnia in
June of 1995, was unheard of by the American
public prior to the incident. If a reader tuned into news on this
event days later, descriptions from the initial articles may be more
useful. A summarizer that has access to different descriptions will be
able to select the description that best suits both the reader and the
series of articles being summarized.

In this paper, we describe a system called PROFILE that tracks prior
references to a given entity  by extracting descriptions for later use in
summarization. In contrast with previous work on information extraction,
our work has the following features: 

\begin{itemize}
\item It builds a database of profiles for entities by storing
descriptions from a collected corpus of past news.
\item It operates in real time, allowing for connections with the
latest breaking, online
news  to extract information about the most recently mentioned
individuals and organizations.
\item It collects and merges information from distributed sources thus allowing
for a more complete record of information.
\item As it parses and identifies descriptions, it builds a
lexicalized, syntactic representation of the description in a form
suitable for input to the FUF/SURGE language generation system
\cite{Elhadad93,robin-phd}. 
\end{itemize}

The result is a system that can combine descriptions from articles
appearing only a few minutes before the ones being summarized with
descriptions from past news in a permanent record for future use.
Its utility lies in its potential for representing  entities, present in
one article, with descriptions found in other articles, possibly coming
from another source.

Since the system constructs a lexicalized, syntactic functional
description (FD) from the extracted description,  the generator can re-use
the description in new contexts, merging it with other 
descriptions, into a new grammatical sentence. This would not be
possible if only canned strings were used, with no information
about their internal structure. Thus, in addition to collecting a
knowledge source which provides identifying features of individuals,
PROFILE also provides a lexicon of domain appropriate phrases that can be
integrated with individual words from a generator's lexicon to
flexibly produce summary wording. 

We have extended the system by semantically categorizing descriptions using
WordNet \cite{Miller&al90}, so that a generator can more easily determine
which description is relevant in different contexts.

PROFILE can also be used in a real-time fashion to monitor entities and the
changes of descriptions associated with them over the course of time.

In the following sections, we first overview related work in the area
of information extraction. We then turn to a discussion of the 
system components which build the profile database, followed by a
description of how the results are used in generation. We close with our
current directions, describing what parameters can influence a strategy for
generating a sequence of anaphoric references to the same entity over time.

%%%%%%%%%%%%%%%%%%%%%%%%%%%%%%%%%%%%%%%%%%%%%%%%%%%%%%%%%%%%%%%%%%%%%%%%%%%
%SECTION 2
%%%%%%%%%%%%%%%%%%%%%%%%%%%%%%%%%%%%%%%%%%%%%%%%%%%%%%%%%%%%%%%%%%%%%%%%%%%
\section{Related Work}

Research related to ours falls into two main categories:
extraction of information from input text and construction of
knowledge sources for generation.

\subsection{Information Extraction}

Work on information extraction is quite broad and covers far more
topics and problems than the information extraction problem we
address. We restrict our comparison here to work on proper noun
extraction, extraction of people descriptions in various information
extraction systems developed for the message understanding conferences
\cite{muc}, and use of extracted information for question answering.

Techniques for proper noun extraction include the use of regular
grammars to delimit and identify proper nouns \cite{Manietal93,Paiketal93},
the use of extensive name lists, place names, titles and 
``gazetteers'' in conjunction with partial grammars in order to
recognize proper nouns as unknown words in close proximity to known words 
\cite{cowieetal92,aberdeenetal92}, statistical training to learn, for
example, Spanish names, from online corpora \cite{Ayusoetal92}, and
the use of concept based pattern matchers that use semantic concepts
as pattern categories as well as part-of-speech information
\cite{Weischedeletal93,Lehnertetal93}. In addition, some researchers
have explored the use of both local context surrounding the
hypothesized proper nouns \cite{McDonald93,Coates-Stephens91} and the
larger discourse context \cite{Manietal93} to improve the accuracy of
proper noun extraction when large known word lists are not
available. Like this research, our work also aims at extracting proper
nouns without the aid of large word lists. We use a regular
grammar encoding part-of-speech categories to extract certain text patterns
and we use WordNet \cite{Miller&al90} to provide semantic filtering.

Our work on extracting descriptions is quite similar to the work
carried out under the DARPA message understanding program for
extracting descriptions \cite{muc}. The purpose for and the scenario in
which description extraction is done is quite different, but the
techniques are very similar. It is based on the paradigm of
representing patterns that express the kinds of descriptions we
expect; unlike previous work we do not encode semantic categories in
the patterns since we want to capture all descriptions regardless of domain.

Research on a system called {\tt Murax} \cite{Kupiec93} is similar to
ours from a different perspective. {\tt Murax} also extracts
information from a text to serve directly in response to a user
question. {\tt Murax} uses lexico-syntactic patterns, collocational
analysis, along with information retrieval statistics, to find the
string of words in a text that is most likely to serve as an answer to
a user's wh-query. In our work, the string that is extracted may be
merged, or regenerated, as part of a larger textual summary.

\subsection{Construction of Knowledge Sources for Generation}

The construction of a database of phrases for re-use in generation is
quite novel. Previous work on extraction of collocations for use in
generation \cite{Smadja&McKeown91} is related in that full phrases are
extracted and syntactically typed so that they can be merged with
individual words in a generation lexicon to produce a full sentence.
However, extracted collocations were used only to determine {\em
realization} of an input concept. In our work, stored phrases would be
used to provide content that can identify a person or place for a
reader, in addition to providing the actual phrasing.

%%%%%%%%%%%%%%%%%%%%%%%%%%%%%%%%%%%%%%%%%%%%%%%%%%%%%%%%%%%%%%%%%%%%%%%%%%%
%SECTION 3
%%%%%%%%%%%%%%%%%%%%%%%%%%%%%%%%%%%%%%%%%%%%%%%%%%%%%%%%%%%%%%%%%%%%%%%%%%%
\section{Creation of a Database of Profiles}

\begin{figure}[htbp]
\center {
\epsfxsize=3.5in
\leavevmode
\epsfbox{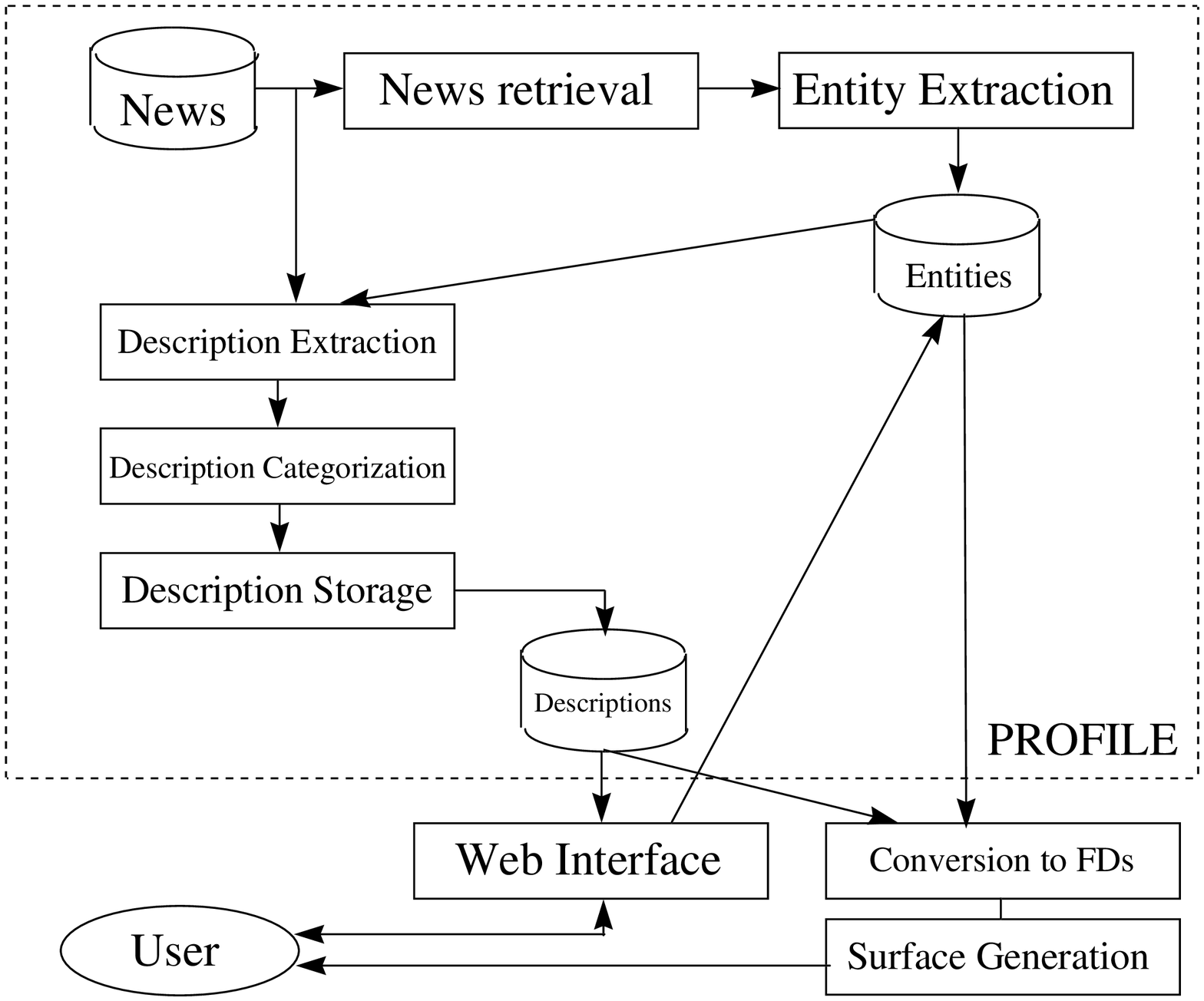}
\caption {Overall Architecture of PROFILE. \label{figure-arch}}
}
\end{figure}

Figure~\ref{figure-arch} shows the overall architecture of PROFILE and the
two interfaces to it (a user interface on the World-Wide Web and an
interface to 
a natural language generation system). In this section, we describe  
the extraction component of PROFILE, the following section focuses on
the uses of PROFILE for generation, and Section~\ref{section-contributions}
describes the Web-based interface.

%%%%%%%%%%%%%%%%%%%%
%  SUBSECTION 3.1  %
%%%%%%%%%%%%%%%%%%%%
\subsection{Extraction of entity names from old newswire}
\label{subsection-3.1}

To seed the database with an initial set of descriptions, we used a 1.4 MB
corpus containing Reuters newswire from March to June of 1995. The purpose
of such an initial set of descriptions is twofold. First, it allows us to
test the other components of the system. Furthermore, at the time a
description is needed it  limits  the amount of online full text, Web
search that must be done. At this stage, search is limited to the database
of retrieved descriptions only, thus reducing search time as no connections
will be made to external news sources at the time of the query. Only when a
suitable stored description cannot be found will the system initiate
search of additional text.

\begin{itemize}
\item {\bf Extraction of candidates for proper nouns}. After tagging the
corpus using the POS part-of-speech tagger \cite{Church88}, we used a CREP
\cite{Duford93} regular grammar to first extract all possible candidates 
for entities. These consist of all sequences of words that were tagged as
proper nouns (NP) by POS. Our manual analysis showed that out of a total of
2150 entities recovered in this way, 1139 (52.9\%) are not names of
entities. Among these are n-grams such as ``Prime Minister'' or ``Egyptian
President'' which were tagged as NP by POS. 
Table~\ref{table-corpus-analysis} shows how many entities we retrieve at
this stage, and of them, how many pass the semantic filtering test. The
numbers in the left-hand column refer to two-word noun phrases that
identify entities (e.g., ``Bill Clinton''). Counts for three-word noun
phrases are shown in the right-hand column. We show counts for multiple and
unique occurrences of the same noun phrase.

\item {\bf Weeding out of false candidates}. Our system analyzed all
candidates for entity names using WordNet \cite{Miller&al90} and removed
from consideration those that contain words appearing in WordNet's
dictionary. This resulted in a list of 421 unique entity names that we used
for the automatic description extraction stage. All 421 entity names
retrieved by the system are indeed proper nouns.

\end{itemize}

\begin{table*}[htbp]
\footnotesize
\centering
\begin{tabular}{|l|l|l|l|l|}\hline
 & \multicolumn{2}{|c|}{{\bf Two-word entities}} &
\multicolumn{2}{|c|}{{\bf Three-word entities}} \\ \cline{2-3}
\cline{4-5} {\bf Stage} & {\bf Entities} & {\bf Unique Entities} & {\bf
Entities} & {\bf Unique Entities} \\ \hline   
POS tagging only & 9079 & 1546 & 2617 & 604 \\ 
After WordNet checkup & 1509 & 395 & 81 & 26 \\ \hline
\end{tabular}
\caption{Two-word and three-word entities retrieved by the system. \label{table-corpus-analysis}}
\end{table*}

%%%%%%%%%%%%%%%%%%%%
%  SUBSECTION 3.2  %
%%%%%%%%%%%%%%%%%%%%
\subsection{Extraction of descriptions}
There are two occasions on which we extract descriptions using finite-state
techniques. The first case is when the entity that we want to describe was
already extracted automatically (see Subsection~\ref{subsection-3.1})
and exists in PROFILE's database. The second
case is when we want a description to be retrieved in real time based on a
request from either a Web user or the generation system.

There exist many live sources of newswire on the Internet that can
be used for this second case. Some that merit
our attention are the ones that can be accessed remotely through small
client programs that don't require any sophisticated protocols to access
the newswire articles. Such sources include HTTP-accessible sites such as
the Reuters site at www.yahoo.com and CNN Interactive at www.cnn.com, as
well as others such as ClariNet which is propagated through the NNTP
protocol. All these sources share a common characteristic in that they are all
updated in real time and all contain information about current events.
Hence, they are therefore likely to satisfy the criteria of pertinence to
our task, such as the likelihood of the sudden appearance of new entities
that couldn't possibly have been included a priori in the generation
lexicon.

Our system generates finite-state representations of the entities that need
to be described. 
An example of a finite-state description of the entity
``Yasser Arafat'' is shown in
Figure~\ref{figure-sample-fsgrammar}. These full expressions are used
as input to the description finding module which uses them to find
candidate sentences in the corpus for finding descriptions. Since the
need for a description may arise at a later time than when the entity
was found and may require searching new text, the description finder
must first locate these expressions in the text.

\begin{figure*}[htbp]
\footnotesize
\centering
\begin{tabular}{|l|} \hline
{\tt SEARCH\_STRING = ((\{NOUN\_PHRASE\}\{SPACE\})+\{SEARCH\_0\})|(\{SEARCH\_0\}\{SPACE\}\{COMMA\}\{SPACE\}\{NOUN\_PHRASE\})} \\
{\tt SEARCH\_109 = [Yy]asser\{T\_NOUN\}\{SPACE\}[Aa]rafat\{T\_NOUN\}} \\
{\tt SEARCH\_0 = \{SEARCH\_1\}|\{SEARCH\_2\}|...|\{SEARCH\_109\}|...} \\ \hline
\end{tabular}
\caption{Finite-state representation of ``Yasser Arafat''. \label{figure-sample-fsgrammar}}
\end{figure*}

These representations are fed to CREP which extracts noun phrases on
either side of the entity (either pre-modifiers or appositions) from
the news corpus.  The finite-state grammar for noun phrases that we
use represents a variety of different syntactic structures for both
pre-modifiers and appositions. Thus, they may range from a simple noun
(e.g., ``president Bill Clinton'') to a much longer expression (e.g.,
``Gilberto Rodriguez Orejuela, the head of the Cali cocaine cartel'').
Other forms of descriptions, such as relative clauses, are the focus
of ongoing implementation.

Table~\ref{table-descriptions} shows some of the different patterns
retrieved. 

\begin{table*}[htbp]
\footnotesize
\centering
\begin{tabular}{|l|l|l|}\hline
{\bf Example} & {\bf Trigger Term} & {\bf Semantic Category} \\ \hline 
Addis Ababa, {\it the Ethiopian capital} & capital & location \\
{\it South Africa's main black opposition leader}, Mangosuthu Buthelezi & leader & occupation \\
Boerge Ousland, {\it 33} & 33 & age \\
{\it maverick French ex-soccer boss} Bernard Tapie & boss & occupation \\
{\it Italy's former prime minister}, Silvio Berlusconi & minister & occupation \\
Sinn Fein, {\it the political arm of the Irish Republican Army} & arm &
organization \\ \hline
\end{tabular}
\caption{Examples of retrieved descriptions. \label{table-descriptions}}
\end{table*}

%%%%%%%%%%%%%%%%%%%%
%  SUBSECTION 3.3  %
%%%%%%%%%%%%%%%%%%%%
\subsection{Categorization of descriptions}

We use WordNet to group extracted descriptions into categories. For all
words in the description, we try to find a WordNet hypernym that can
restrict the semantics of the description. Currently, we identify concepts
such as ``profession'', ``nationality'', and ``organization''. Each of
these concepts is triggered by one or more words (which we call
``triggers'') in the description. Table~\ref{table-descriptions} shows some
examples of descriptions and the concepts under which they are classified
based on the WordNet hypernyms for some ``trigger'' words. For example, all
of the following ``triggers'' in the list ``minister'', ``head'',
``administrator'', and ``commissioner'' can be traced up to ``leader''
in the WordNet hierarchy.

%%%%%%%%%%%%%%%%%%%%
%  SUBSECTION 3.4  %
%%%%%%%%%%%%%%%%%%%%
\subsection{Organization of descriptions in a database of profiles}

For each retrieved entity we create a new profile in a database of
profiles. We keep information about the surface string that is used to
describe the entity in newswire (e.g., ``Addis Ababa''), the source of the
description and the date that the entry has been made in the database
(e.g., ``reuters95\_06\_25''). In addition to these pieces of
meta-information, all retrieved descriptions and their frequencies are also
stored.

Currently, our system doesn't have the capability of matching references to
the same entity that use different wordings. As a result, we keep separate
profiles for each of the following: ``Robert Dole'', ``Dole'', and ``Bob
Dole''. We use each of these strings as the key in the database of
descriptions. 

Figure ~\ref{figure-major-profile} shows the profile
associated with the key ``John Major''.

\begin{figure}[htbp]
\footnotesize
\centering
\begin{tabular}{|l|} \hline
{\tt KEY: john major} \\
{\tt SOURCE: reuters95\_03-06\_.nws} \\
{\tt DESCRIPTION: british prime minister} \\
{\tt FREQUENCY: 75} \\
{\tt DESCRIPTION: prime minister} \\
{\tt FREQUENCY: 58} \\
{\tt DESCRIPTION: a defiant british prime minister} \\
{\tt FREQUENCY: 2} \\
{\tt DESCRIPTION: his british counterpart} \\
{\tt FREQUENCY: 1} \\ \hline
\end{tabular}
\caption{Profile for John Major. \label{figure-major-profile}}
\end{figure}

The database of profiles is updated every time a query retrieves new
descriptions matching a certain key.

%%%%%%%%%%%%%%%%%%%%%%%%%%%%%%%%%%%%%%%%%%%%%%%%%%%%%%%%%%%%%%%%%%%%%%%%%%%
%SECTION 4
%%%%%%%%%%%%%%%%%%%%%%%%%%%%%%%%%%%%%%%%%%%%%%%%%%%%%%%%%%%%%%%%%%%%%%%%%%%
\section{Generation}

We have made an attempt to reuse the descriptions, retrieved by the system,
in more than a trivial way. The content planner of a language generation
system that needs to present an entity to the user that he has not seen
previously, might want to include some background information about
it. However, in case the extracted information doesn't contain a handy
description, the system can use some descriptions retrieved by PROFILE.

\subsection{Transformation of descriptions into Functional Descriptions}

Since our major goal in extracting descriptions from on-line corpora was to
use them in generation, we have written a utility which converts
finite-state descriptions retrieved by PROFILE into functional descriptions
(FD) \cite{Elhadad88} that we can use directly in generation. A description
retrieved by the system from the article in~\ref{fig:source-article} is
shown in Figure~\ref{figure-silvio-description}. The corresponding FD is
shown in Figure~\ref{figure-silvio-fd}.  

\begin{figure}[t]
MILAN - A judge ordered Italy's former prime minister Silvio
Berlusconi to stand trial in January on corruption charges in a
ruling that could destroy the media magnate's hope of returning
to high office. 
\caption{Source article.}
\label{fig:source-article} 
\end{figure}

\begin{figure}[htbp]
\footnotesize
\centering
\begin{tabular}{|l|} \hline
{\tt Italy@NPNP 's@\$ former@JJ prime@JJ} \\
{\tt minister@NN Silvio@NPNP Berlusconi@NPNP} \\ \hline
\end{tabular}
\caption{Retrieved description for Silvio Berlusconi. \label{figure-silvio-description}}
\end{figure}

\begin{figure}[htbp]
\scriptsize
\centering
\begin{verbatim}
    ((cat np)
     (complex apposition)
     (restrictive no)
     (distinct ~(((cat common)
                  (possessor ((cat common)
                              (determiner none)
                              (lex ``Italy'')))
                  (classifier ((cat noun-compound)
                               (classifier ((lex ``former'')))
                               (head ((lex ``prime'')))))
                  (head ((lex ``minister''))))
                 ((cat person-name)
                  (first-name ((lex ``Silvio'')))
                  (last-name ((lex ``Berlusconi''))))))))
\end{verbatim}
\caption{Generated FD for Silvio Berlusconi. \label{figure-silvio-fd}}
\end{figure}

We have implemented a TCP/IP interface to Surge. The FD generation
component uses this interface to send a new FD to the surface realization
component of Surge which generates an English surface form corresponding to
it.  

\subsection{Lexicon creation}

We have identified several major advantages of using FDs produced by the
system in generation compared to using canned phrases.

\begin{itemize}
\item {\bf Grammaticality}. The deeper representation allows for
grammatical transformations, such as aggregation: e.g., ``president
Yeltsin'' $+$ ``president Clinton'' can be generated as ``presidents
Yeltsin and Clinton''. 

\item {\bf Unification with existing ontologies}. E.g., if an ontology
contains information about the word ``president'' as being a realization of
the concept ``head of state'', then under certain conditions, the
description can be replaced by one referring to ``head of state''.

\item {\bf Generation of referring expressions}. In the previous example, if
``president Bill Clinton'' is used in a sentence, then ``head of state''
can be used as a referring expression in a subsequent sentence.

\item {\bf Enhancement of descriptions}. If we have retrieved ``prime
minister'' as a description for Silvio Berlusconi, and later we obtain
knowledge that someone else has become Italy's primer minister, then we can
generate ``former prime minister'' using a transformation of the old FD.

\item {\bf Lexical choice}. When different descriptions are automatically
marked for semantics, PROFILE can prefer to generate one over another based
on semantic features.  This is useful if a summary discusses events
related to one description associated with the entity more than the
others. 

\item {\bf Merging lexicons}. The lexicon generated automatically by the
system can be merged with a domain lexicon generated manually.
\end{itemize}

These advantages look very promising and we will be exploring
them in detail in our work on summarization in the near future.

%%%%%%%%%%%%%%%%%%%%%%%%%%%%%%%%%%%%%%%%%%%%%%%%%%%%%%%%%%%%%%%%%%%%%%%%%%%
%SECTION 5
%%%%%%%%%%%%%%%%%%%%%%%%%%%%%%%%%%%%%%%%%%%%%%%%%%%%%%%%%%%%%%%%%%%%%%%%%%%
\section{Coverage and Limitations}

In this section we provide an analysis of the capabilities and
current limitations of PROFILE.

\subsection{Coverage}

At the current stage of implementation, PROFILE has the following coverage.

\begin{itemize}
\item {\bf Syntactic coverage}. Currently, the system includes an extensive
finite-state grammar that can handle various pre-modifiers and appositions. 
The grammar matches arbitrary noun phrases in each of these two cases to
the extent that the POS part-of-speech tagger provides a correct tagging.

\item {\bf Precision}. In Subsection~\ref{subsection-3.1} we showed the
precision of the extraction of entity names. Similarly, we have computed
the precision of retrieved 611 descriptions using randomly selected
entities from the list retrieved in Subsection~\ref{subsection-3.1}. Of the
611 descriptions, 551 (90.2\%) were correct. The others included a roughly
equal number of cases of incorrect NP attachment and incorrect
part-of-speech assignment. For our task (symbolic text generation),
precision is more important than recall; it is critical that the extracted
descriptions are correct in order to be converted to FD and generated.

\item {\bf Length of descriptions}. The longest description retrieved by the
system was 9 lexical items long: ``Maurizio Gucci, the former head of
Italy's Gucci fashion dynasty''. The shortest descriptions are 1 lexical
item in length - e.g. ``President Bill Clinton''.

\item {\bf Protocol coverage}. We have implemented retrieval facilities to
extract descriptions using the NNTP (Usenet News) and HTTP (World-Wide Web)
protocols. 
\end{itemize}

\subsection{Limitations}

Our system currently doesn't handle entity cross-referencing. It will not
realize that ``Clinton'' and ``Bill Clinton'' refer to the same person. Nor
will it link a person's profile with the profile of the organization of
which he is a member. 

At this stage, the system generates functional descriptions (FD), but they
are not being used in a summarization system yet.

%%%%%%%%%%%%%%%%%%%%%%%%%%%%%%%%%%%%%%%%%%%%%%%%%%%%%%%%%%%%%%%%%%%%%%%%%%%
%SECTION 6
%%%%%%%%%%%%%%%%%%%%%%%%%%%%%%%%%%%%%%%%%%%%%%%%%%%%%%%%%%%%%%%%%%%%%%%%%%%
\section{Current Directions}

One of the more important current goals is to increase coverage of the
system by providing interfaces to a large number of on-line sources of
news. We would ideally want to build a comprehensive and shareable database
of profiles that can be queried over the World-Wide Web.

We need to refine the algorithm to handle cases that are currently
problematic. For example, polysemy is not properly handled. For instance,
we would not label properly noun phrases such as ``Rice University'', as it
contains the word ``rice'' which can be categorized as a food. 

Another long-term goal of our research is the generation of evolving
summaries 
that continuously update the user on a given topic of interest. In that
case, the system will have a model containing all prior interaction
with the user. To avoid repetitiveness, such a system will have to resort to
using different descriptions (as well as referring expressions) to address
a specific entity\footnote{Our corpus analysis supports this proposition
-- a large number of threads of summaries on the same topic from the
Reuters and UPI newswire used up to 10 different referring expressions
(mostly of the type of descriptions discussed in this paper) to refer to
the same entity.}. We will be investigating an algorithm that will select a
proper ordering of multiple descriptions referring to the same person.

After we collect a series of descriptions for each possible entity, we need
to decide how to select among all of them. There are two scenarios. In
the first one, we have to pick one single description from the database
which best fits the summary that we are generating. In the second scenario,
the evolving summary, we have to generate a {\it sequence} of descriptions,
which might possibly view the entity from different perspectives. We are
investigating algorithms that will decide the order of generation of the
different descriptions. Among the factors that will influence the selection
and ordering of descriptions, we can note the user's interests, his
knowledge of the entity, the focus of the summary (e.g., ``democratic
presidential candidate'' for Bill Clinton, vs. ``U.S. president''). We can
also select one description over another based on how recent they have been
included in the database, whether or not one of them has been used in a
summary already, whether the summary is an update to an earlier summary,
and whether another description from the same category has been used
already. We have yet to decide under what circumstances a description needs
to be generated at all. 

We are interested in implementing existing algorithms or designing our own
that will match different instances of the same entity appearing in
different syntactic forms - e.g., to establish that ``PLO'' is an alias for
the ``Palestine Liberation Organization''. We will investigate using
cooccurrence information to match acronyms to full organization names and
alternative spellings of the same name with each other.

An important application that we are considering is applying the technology
to text available using other protocols - such as SMTP (for electronic
mail) and retrieve descriptions for entities mentioned in such messages.

We will also look into connecting the current interface with news available
to the Internet with an existing search engine such as Lycos
(www.\-lycos.\-com) or AltaVista (www.\-altavista.\-digital.\-com). We can
then use the existing indices of all Web documents mentioning a given
entity as a news corpus on which to perform the extraction of
descriptions. 

Finally, we will investigate the creation of KQML \cite{Finin&al94}
interfaces to the different components of PROFILE which will be linked to
other information access modules at Columbia University.

%%%%%%%%%%%%%%%%%%%%%%%%%%%%%%%%%%%%%%%%%%%%%%%%%%%%%%%%%%%%%%%%%%%%%%%%%%%
%SECTION 7
%%%%%%%%%%%%%%%%%%%%%%%%%%%%%%%%%%%%%%%%%%%%%%%%%%%%%%%%%%%%%%%%%%%%%%%%%%%

\section{Contributions}
\label{section-contributions}

\begin{figure}[htbp]
\center {
\epsfxsize=3.2in
\leavevmode
\epsfbox{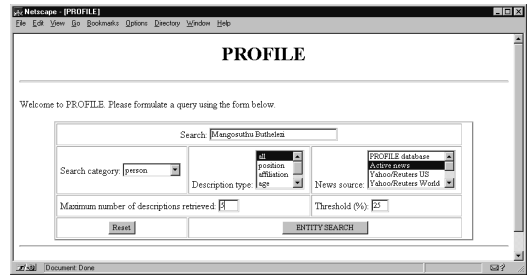}
\caption {Web-based interface to PROFILE. \label{figure-web-interface}}
}
\end{figure}

We have described a system that allows users to retrieve descriptions of
entities using a Web-based search engine. Figure~\ref{figure-web-interface}
shows the Web interface to PROFILE. Users can select an entity (such as
``John Major''), specify what semantic classes of descriptions they want to
retrieve (e.g., age, position, nationality) as well as the maximal number
of queries that they want. They can also specify which sources of
news should be searched. Currently, the system has an interface to Reuters
at www.yahoo.com, The CNN Web site, and to all local news delivered via
NNTP to our local news domain.

The Web-based interface is accessible publicly (currently within Columbia
University only). All queries are cached and the descriptions retrieved can
be reused in a subsequent query. We believe that such an approach to
information extraction can be classified as a collaborative database.

The FD generation component produces syntactically correct functional
descriptions that can be used to generate English-language descriptions
using FUF and Surge, and can also be used in a general-purpose
summarization system in the domain of current news.

All components of the system assume no prior domain knowledge and are
therefore portable to many domains - such as sports, entertainment, and	
business.

\section{Acknowledgments}

This work was partially  supported by   NSF grant GER-90-2406 and by 
a grant from Columbia University's Strategic Initiative Fund sponsored by 
the Provost's Office.

%\bibliography{/u/bet/radev/paper/1997/lexicon/anlp97}
%\bibliographystyle{/u/bet/radev/paper/styles/acl}

\end{document}